\documentclass[a4paper,fleqn]{cas-dc}

\usepackage[numbers,sort&compress]{natbib}

\def\tsc#1{\csdef{#1}{\textsc{\lowercase{#1}}\xspace}}
\tsc{WGM}
\tsc{QE}

\usepackage{algorithm}
\usepackage{algorithmic}
\usepackage{multirow}
\usepackage{booktabs}
\usepackage[switch]{lineno} 
\usepackage{amssymb}
\usepackage{amsmath}
\usepackage{bm}
\usepackage{float}
\usepackage{graphicx}
\usepackage{color}
\usepackage{makecell}

\begin{document}
\let\WriteBookmarks\relax
\def\floatpagepagefraction{1}
\def\textpagefraction{.001}

\newcommand{\todo}[1]{{\color{black}#1}}
\newcommand{\cjy}[1]{{\color{black}#1}}
\newcommand{\jy}[1]{{\color{black}#1}}
\newcommand{\err}[1]{{\color{black}#1}}

\shorttitle{Embracing Ambiguity: Improving Similarity-oriented Tasks with Contextual Synonym Knowledge}    

\shortauthors{Li et al.}  

\title [mode = title]{Embracing Ambiguity: Improving Similarity-oriented Tasks with Contextual Synonym Knowledge}  



%


\author[1]{Yangning Li}[
      orcid=0000-0002-1991-6698,
    ]



\ead{liyn20@mails.tsinghua.edu.cn}


\credit{Conceptualization of this study, Methodology, Experiments}

\affiliation[1]{organization={Shenzhen International Graduate School, Tsinghua University},
            city={Shenzhen},
            postcode={518055}, 
            state={Guangdong},
            country={China}}

\author[2]{Jiaoyan Chen}[]
\credit{Conceptualization of this study, Methodology, Experiments}
\ead{jiaoyan.chen@manchester.ac.uk}
\author[1]{Yinghui Li}[]
\credit{Investigation process, Experimental verification}
\author[1]{Tianyu Yu}[]
\credit{Investigation process, Experimental verification}
\author[3]{Xi Chen}[]
\credit{Revision of the paper, Funding Support}
\author[1,4,]{Hai-Tao Zheng}[]
\cormark[1]
\credit{Revision of the paper, Funding Support}
\ead{zheng.haitao@sz.tsinghua.edu.cn}




\affiliation[2]{organization={Department of Computer Science, The University of Manchester},
            city={Manchester},
            postcode={M13 9PL}, 
            country={UK}}

\affiliation[3]{organization={Platform and Content Group, Tencent},
            city={Shenzhen},
            postcode={518055}, 
            state={Guangdong},
            country={China}}

\affiliation[4]{organization={Peng Cheng Laboratory},
            city={Shenzhen},
            postcode={518055}, 
            state={Guangdong},
            country={China}}
            
\cortext[1]{Corresponding author}



\begin{abstract}
Contextual synonym knowledge is crucial for 
\cjy{those} similarity-oriented tasks whose core challenge lies in capturing semantic similarity between entities in \cjy{their} contexts, such as entity linking and entity matching. 
However, most Pre-trained Language Models (PLMs) lack synonym knowledge due to inherent limitations of \cjy{their pre-training objectives such as} masked language modeling (MLM). 
Existing works which 
\cjy{inject}
synonym knowledge into PLMs \cjy{often} suffer from two severe problems: 
\textit{(i)} \jy{Neglecting the ambiguity of synonyms, 
} and \textit{(ii)} Undermining semantic understanding of original PLMs, which \jy{is} caused by inconsistency between the \textit{exact semantic similarity} of the synonyms and the \textit{broad conceptual relevance} learned from the original corpus.
To address these issues, we propose PICSO, a flexible framework that supports the injection of contextual synonym knowledge from multiple domains into PLMs via a novel entity-aware Adapter which focus\jy{es} on the semantics of the entities (synonyms) in the contexts. Meanwhile, PICSO store\jy{s} the synonym knowledge in additional parameters of the Adapter structure, \jy{which prevents it from corrupting}
the semantic understanding
of the original PLM.
Extensive experiments demonstrate that PICSO \cjy{can dramatically outperform the original PLMs and the other knowledge and synonym injection models on four different} similarity-oriented tasks. In addition, experiments on GLUE prove that 
PICSO \cjy{also benefits} general natural language understanding tasks. Codes and data will be public.
\end{abstract}


\begin{highlights}
\item Contextual synonym knowledge is extremely effective for similarity-oriented tasks, and we are the first work to inject \textbf{contextual} synonym knowledge into the Pre-trained Language Model (PLM).
\item We propose PICSO, a flexible framework equipped with our designed entity-aware Adapter. PICSO supports continuous injection of synonym knowledge from multiple domains, while the contextual semantic understanding capability of the original PLM is not undermined. 
\item PICSO can dramatically outperform the original PLMs and the other knowledge and synonym injection models on various similarity-oriented tasks. In addition, PICSO also benefits general natural language understanding tasks.
\end{highlights}


\maketitle

\section{Introduction}
Pre-trained language models (PLMs) such as BERT~\citep{kenton2019bert}, RoBERTa~\citep{liu2019roberta} \cjy{and GPT~\citep{radford2018improving} have achieved great success in natural language processing (NLP) due to their semantic understanding capabilities achieved by 
pre-training on large-scale corpora}. However, most PLMs only acquire \cjy{statistical word co-occurrence knowledge through their pre-training objectives such as masked language modeling (MLM) \citep{li2020sentence}, 
which leads to limited capabilities in understanding synonyms.}

\begin{figure}[t]
\centering
\includegraphics[width=0.4\textwidth]{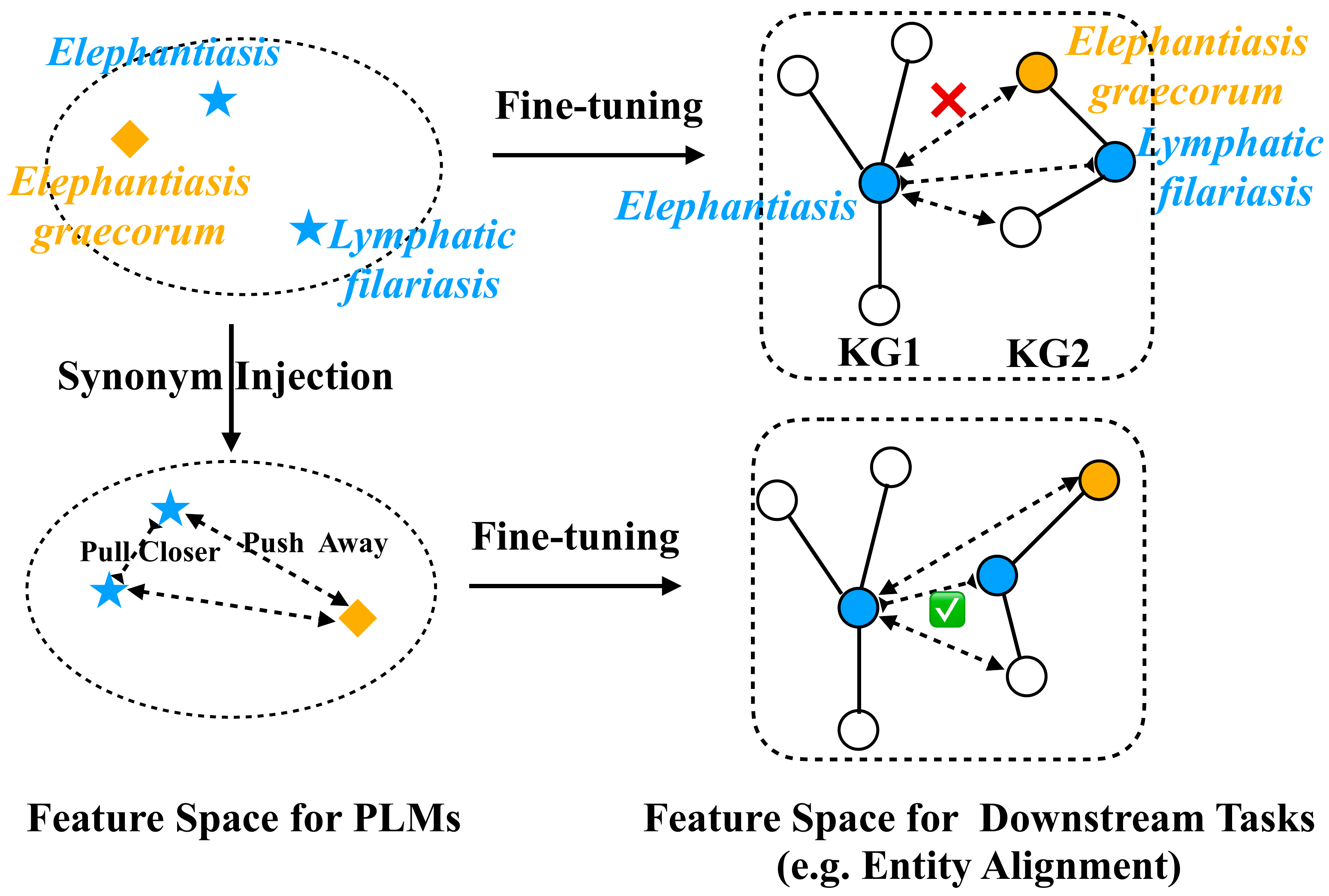}
\caption{An example demonstrating the importance of synonym knowledge for entity alignment. Some normal PLM such as BERT tend to encode entities with common tokens to more similar spaces ignoring synonymic semantics, such as \textit{Elephantiasis} for \textit{Elephantiasis graecorum} and \textit{Lymphatic filariasis} (6.62 vs 9.17 we measured in BERT feature space), which causes misalignment.}
\label{fig:intro}
\end{figure}

Synonym knowledge facilitates models to capture fine-grained semantic relations and is crucial \cjy{in NLP 
especially for addressing \textbf{similarity-oriented tasks}, such as entity linking \citep{zhang2010entity} and} entity resolution \citep{xu2019towards}. The core challenge of such tasks lies in modeling the semantic similarity of entities in complex contexts, \jy{where understanding the synonymous relationship between phrases is essential.}
Taking ontology alignment as an example, \cjy{cross-ontology class pairs with synonymic relationships account for 51\% of the total class mappings in the widely used benchmark FMA-SNOMED of OAEI LargeBio Track\footnote{https://www.cs.ox.ac.uk/isg/projects/SEALS/oaei/}}. 
In \cjy{knowledge graph (KG)} canonicalization, which aims to cluster semantically identical entities, \cjy{about 30\% identical entities} in the Reverb45k dataset \citep{vashishth2018cesi} appear in the \cjy{synonym sets (synsets) of}  UMLS\footnote{Unified Medical Language System (UMLS) is a comprehensive collection of biomedical terms.}.
\cjy{In specific domains, making sense of synonym knowledge could become even more important and challenging. 
\jy{As the biomedical example in Figure \ref{fig:intro} shows, } \textit{Elephantiasis} is synonymous to \textit{Lymphatic filariasis} but non-synonymous to \textit{Elephantiasis graecorum}, although the latter has a closer surface form and would be regarded as synonymous by a normal PLM.}
\cjy{To better address such similarity-oriented tasks, capturing \jy{the synonymous relationship}
between phrases under complex contexts is urgently required}. 
This motivates us to inject synonym knowledge into PLMs.

Some pioneering works have explored injecting synonym knowledge into PLMs. LIBERT \citep{lauscher2020specializing} train\cjy{s} BERT from scratch with \cjy{an auxiliary task that binary classifies whether entity pairs are synonymous pairs}. SAPBERT \citep{liu2021self} pre-train\cjy{s BERT with synsets from UMLS. A metric learning objective is used to optimize the BERT, with synonymous and non-synonymous entity pairs as positive and negative training samples, respectively. 
%
Although promising results have been achieved, these works still suffer from the following two} problems: 

\noindent\textbf{Neglecting the ambiguity of synonyms.}
Synonyms are naturally context-sensitive. \jy{It is} intuitive that some entities are synonyms and thus close to each other in the semantic space, but are comparably different in some specific aspects and thus far away from each other in the corresponding semantic spaces. However, pre-training objectives of the current synonym injection works completely \jy{ignore}
this ambiguity and rigidly pull synonym pairs closer together in semantic space. Take the example in the upper right of Figure \ref{fig:method}, it is unreasonable to always treat \textit{Washington} as synonymous to either \textit{George Washington} or \textit{Washington, D.C.} since \textit{Washington} has different meanings in different contexts. The pre-training objective of closing the distance from \textit{Washington} to \textit{George Washington} and \textit{Washington, D.C.} cannot be satisfied simultaneously. Therefore, Context is imperative to disambiguate synonyms and should be considered in injecting synonym knowledge. Ignoring the synonym context will also significantly limit the generality of the synonym injected PLM, i.e., the PLM is hard to be applied to tasks out of the domain where the training synonyms are extracted.
 
\noindent\textbf{Undermining semantic understanding of original PLMs.} The \textit{exact semantic similarity} expressed by synonyms and the \textit{broad conceptual relevance} implied by MLM are \cjy{often} contradictory \citep{lauscher2020specializing}. 
 \cjy{This is because PLMs that adopt MLM for pre-training (e.g., BERT)}
 acquire semantic understanding capabilities based on word co-occurrence \cjy{statistics}.
 The neighboring words in the feature space of such PLMs are related words rather than synonyms, e.g., the top 10 nearest neighbors of \textit{good} in BERT contain antonyms like \textit{bad}.
 The existing methods \jy{directly inject}
 synonym knowledge on top of the parameters of the PLMs that have established semantic understanding. \jy{This will inevitably result in semantic conflicts and weakens}
 the PLM's original semantic understanding capabilities. We refer to this phenomenon as \textbf{semantic forgetting}, analogous to the catastrophic forgetting \citep{kirkpatrick2017overcoming,kemker2018measuring} of old samples in continual learning community. In another word, we argue that existing work utilizing synonyms for pre-training is task-specific pre-training at the expense of semantic understanding. SAPBERT, for example, is a further pre-training of PLM with \textit{context-free} synonym knowledge for \textit{context-free} entity linking, which sacrifices the ability to understand semantics (as evidenced by the general degradation of performance for various downstream tasks in Section \ref{sec:exp}) and thus is not generalized.
 
 To address these issues, we \cjy{propose} a \textbf{P}re-trained language model \textbf{I}njected with \textbf{C}ontextual \textbf{S}ynonyms kn\textbf{O}wledge (\textbf{PICSO}), whose input \cjy{for pre-training is sentences with marked synonyms rather than synonyms without contexts. In order for PICSO to not only capture the semantics of the entire sentence, but also to focus on the semantics of the entity (synonym) in the context, we \jy{develop a new entity-aware Adapter structure}
 with a novel masked self-attention mechanism. 
 Equipped \jy{with Adapters of such a structure, PICSO supports continuous injection of synonym knowledge from multiple domains, while the contextual semantic understanding capability of the original PLM is not undermined.}
 In the evaluation, we consider a general domain with 12.8 million synonym pairs extracted from Wikidata and a biomedical domain with 3.7 million synonym pairs extracted from UMLS.
Extensive experiments on four similarity-oriented tasks have demonstrated} that PICSO can dramatically outperform the original PLMs and the other knowledge and synonym injection methods including LIBERT and SAPBERT. 
In addition, experiments on GLUE \cjy{have proven} that PICSO also benefits general Natural Language Understanding (NLU) tasks.

\section{Related Work}
\subsection{Injecting Structure Knowledge into PLMs}
\cjy{Despite great success in many NLP tasks}, some works \citep{poerner2019bert, petroni2019language, kassner2020negated} expose that PLMs such as BERT struggle to acquire rich knowledge during pre-training. Some efforts have been made to \cjy{inject structure knowledge into PLMs, which can be divided into three main categories.}
\cjy{The first category is \textbf{KG Injection}}. With the emergence of plenty of general and domain-specific KGs, they have become \cjy{one of the most important knowledge sources}. A body of mainstream research \citep{peters2019knowledge,sun2020colake,wang2021kepler} is devoted to \cjy{inject KG triples (facts) in form of \textit{(subject, relation, object)} into PLMs.}
ERNIE-THU \citep{zhang2019ernie} encodes the entities and relations in Wikidata by \cjy{the KG embedding model TransE}, then integrates entity representations 
based on the alignments between entity mentions and KG entities. K-Adapter \citep{wang2021k} leaves the original \cjy{parameters of the PLM} unchanged and exports representations for structure knowledge \cjy{of the KG}. This is achieved via an additional compact neural model termed Adapter. Note that we also use Adapter, but our work differs from K-Adapter in two fundamental ways: (1) We focus on the gain of clean and efficient synonym knowledge for similarity-oriented tasks. (2) To inject contextual synonym knowledge more precisely, we specially designed entity-aware Adapter, which proved to be particularly effective in Section \ref{exp:abla_entity_aware}.
\cjy{The second category is \textbf{Rule Injection}}. Rules exist as informal constraints or \cjy{logical expressions}, which can import sound explanatory \citep{deng2014large} or precise reasoning capabilities \citep{amizadeh2020neuro} for PLMs. 
\cjy{The last category is \textbf{Syntax-tree Injection}}. Syntactic knowledge guides PLMs to understand the core constituents in sentences.
\cjy{Some studies \citep{bai2021syntax, zhou2020limit} have considered syntax-trees.}

\cjy{Structure knowledge may benefit similarity-oriented tasks to some extent but the synonym knowledge they contain are implicit and only take a small ratio. 
Meanwhile, existing works \citep{zhang2021drop, liu2020k} demonstrate the presence of redundant and irrelevant structure knowledge injected, which may instead lead to negative impact in solving downstream tasks.} 
Compared to pure synonym knowledge, \cjy{structure} knowledge is inefficient and prone to introduce noisy knowledge. 

\begin{table*}[]
\centering
\caption{Taxonomy of entity-level similarity-oriented tasks.}
\begin{tabular}{llll}
\toprule
Source Entity & Target Entity & Typical Tasks & Examples \\ \midrule
 Unstructured Text  & Unstructured Text          & \shortstack{Entity Resolution \\ Lexical Simplification }       &                  \begin{minipage}[b]{0.5\columnwidth}
        \centering
        \raisebox{-.5\height}{\includegraphics[width=\linewidth]{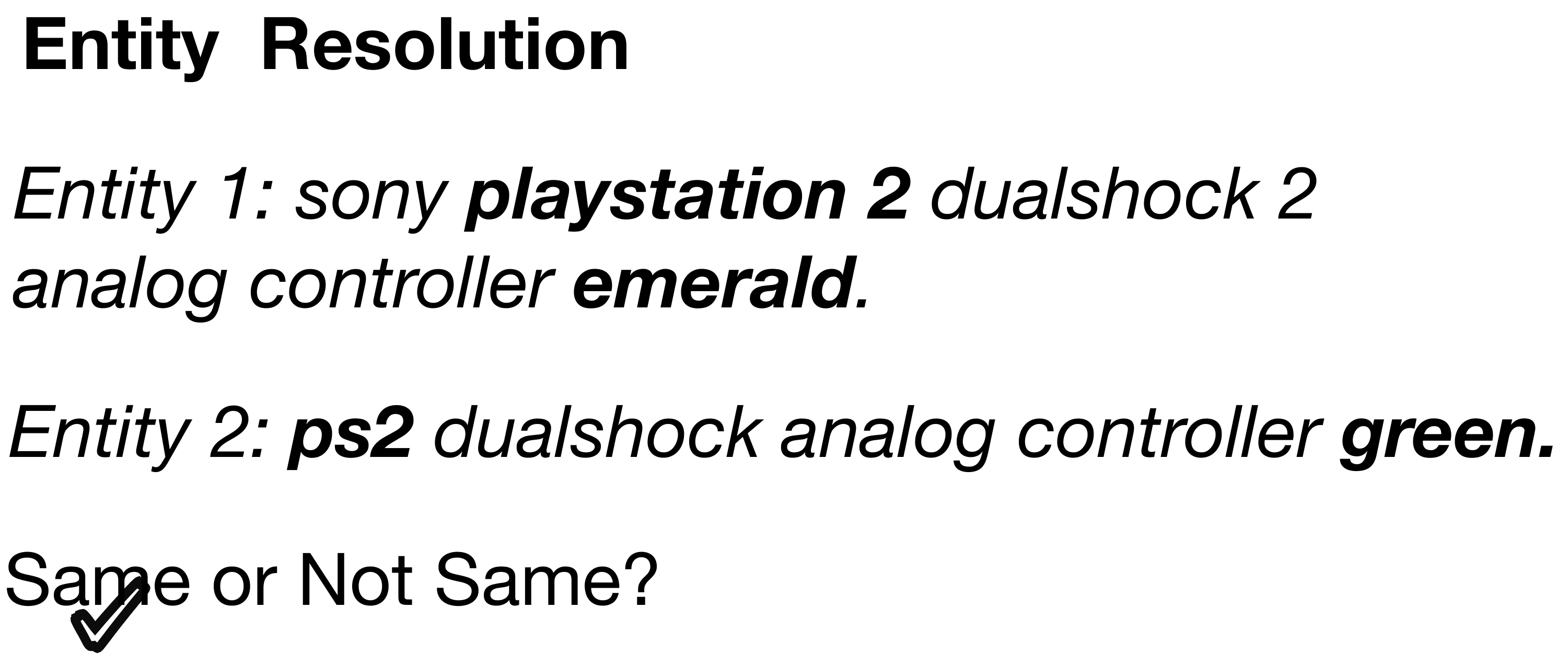}}
    \end{minipage}      \\ \midrule
 Unstructured Text            & Structured KG            & \shortstack{Entity Linking \\ Coreference Resolution}            & \begin{minipage}[b]{0.5\columnwidth}
        \centering
        \raisebox{-.5\height}{\includegraphics[width=\linewidth]{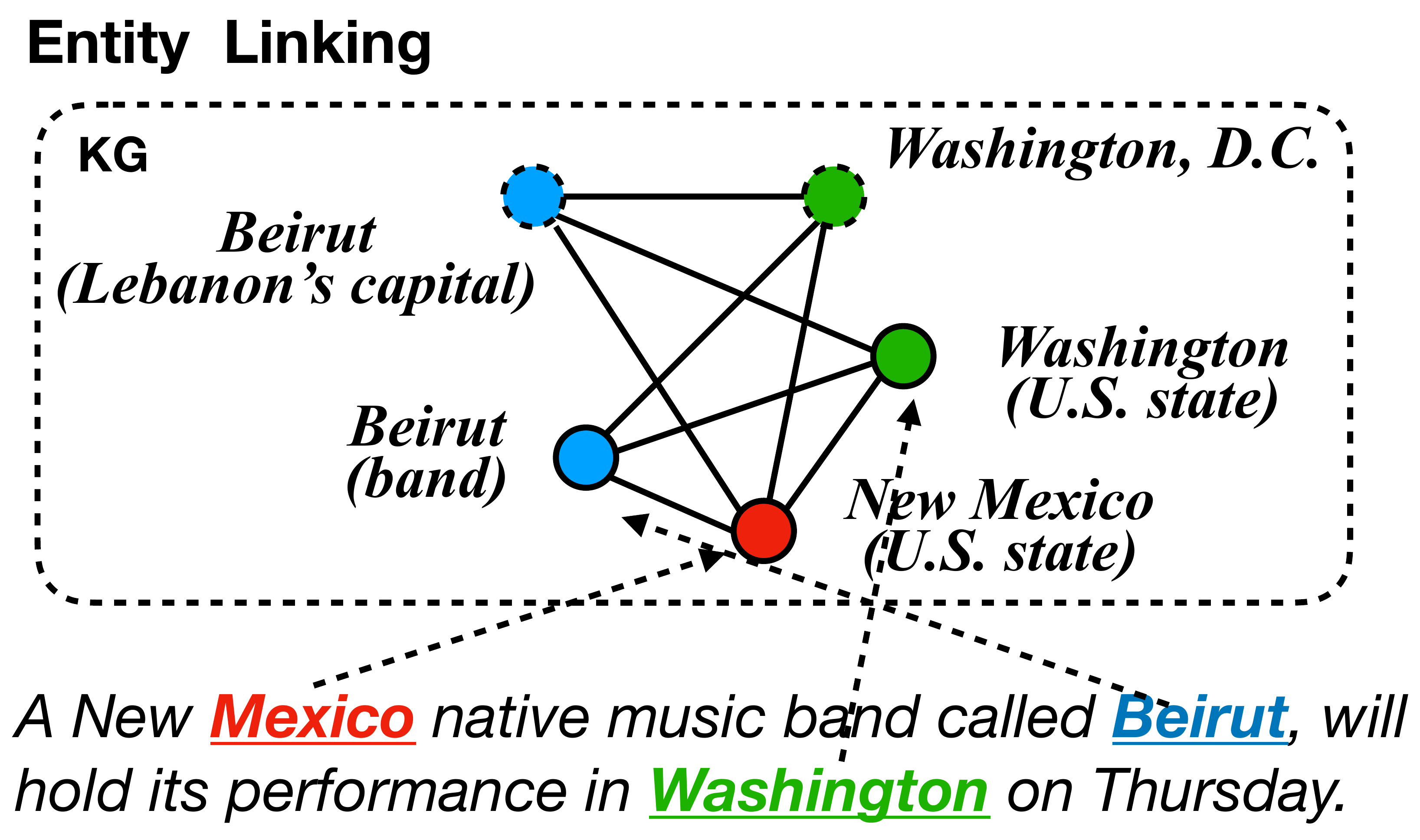}}
    \end{minipage}       \\ \midrule
 Structured KG            & Structured KG            & \shortstack{KG Canonicalization \\ Entity Alignment \\ Ontology Alignment}            & \begin{minipage}[b]{0.5\columnwidth}
        \centering
        \raisebox{-.5\height}{\includegraphics[width=\linewidth]{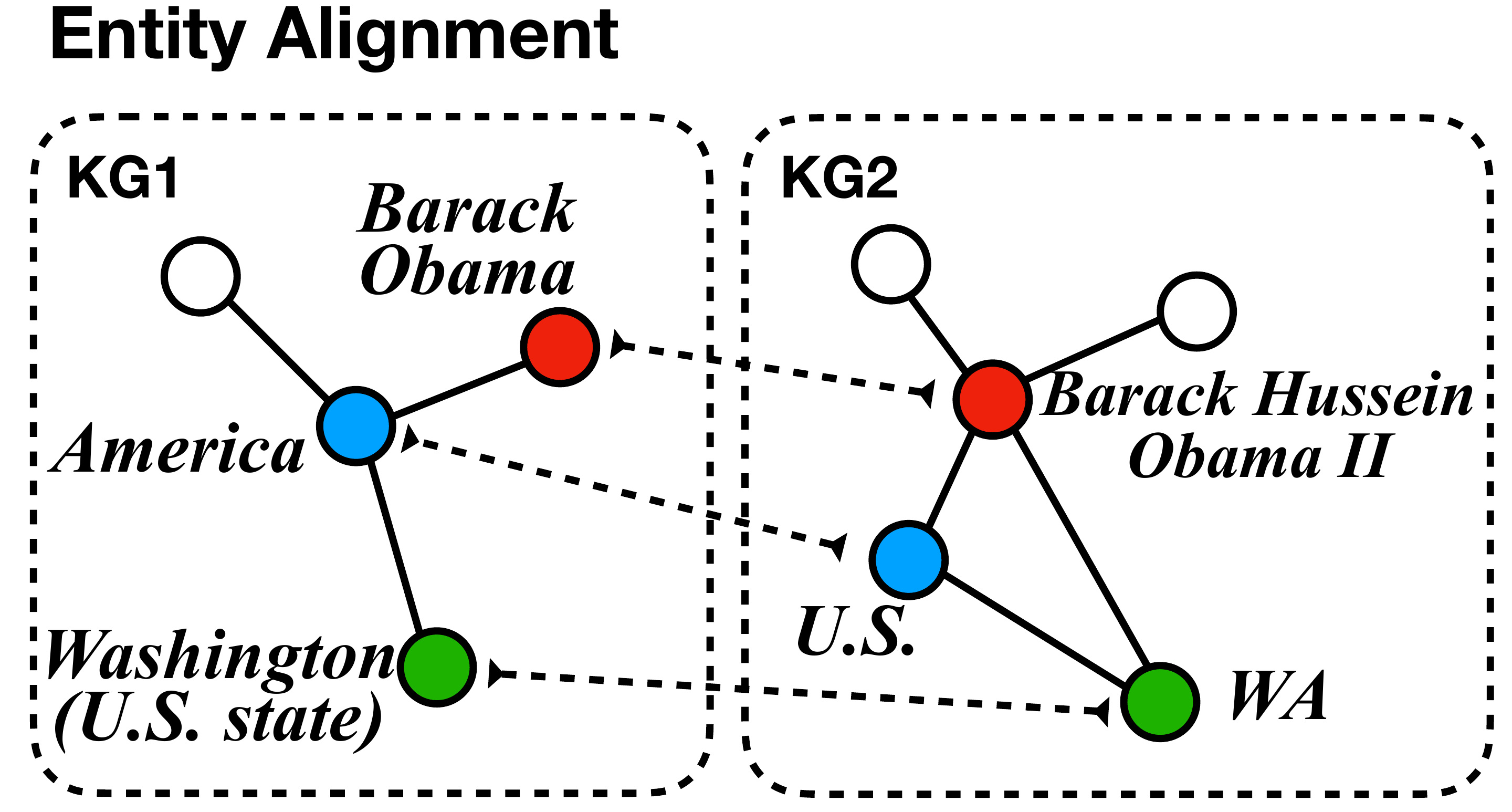}}
    \end{minipage}    \\ \bottomrule
\end{tabular}
\label{tab:rw_entity}
\end{table*}

\subsection{Injecting Synonym Knowledge into PLMs}

The injection of semantic constraints \cjy{from} synonyms into static word representations has been extensively studied before \cjy{PLMs become popular}. Numerous works \citep{mrkvsic2017semantic, glavavs2018explicit} demonstrated that synonyms can help models clearly distinguish between \textit{exact semantic similarity and broader conceptual relatedness}.
These works mainly fall into two categories: \cjy{(1) \textbf{Joint Optimization Models} \citep{nguyen2017hierarchical,osborne2016encoding} which introduce auxiliary objectives in pre-training to constrain the embeddings, and (2) \textbf{Post-optimization Models} \citep{vulic2017cross, ferret2018using} which tune the pre-trained embeddings by adapting the pairwise distances to the semantic constraints of synonyms.}

Following the \cjy{popularity of PLMs, injecting synonym knowledge into PLMs has attracted wide} attention. Some earlier works \citep{tang2021bert,cui2017kbqa} injected task/domain-specific synonym knowledge into PLMs \cjy{during fine-tuning.}
For example, \cjy{BERTMap} \citep{he2022bertmap} collects synonyms in ontologies 
to construct \cjy{corpora which are used to fine-tune BERT for predicting class mappings between ontologies.}
%
Some previous \cjy{works inject synonym knowledge into PLMs during} pre-training. LIBERT \citep{lauscher2020specializing} train\cjy{s} BERT from scratch with \cjy{an auxiliary} binary classification task that predicts whether entity pairs are synonymous or not. SAPBERT \citep{liu2021self} further pre-train\cjy{s} BERT with massive synsets \cjy{extracted from} UMLS. A metric learning objective is used to optimize the BERT, \cjy{where synonymic and non-synonymic pairs are extracted as positive and negative samples, respectively. 
As we have stated before, these methods suffer from ambiguity of synonyms and the semantic forgetting problem. 
Besides, LIBERT has to combine the synonym corpus and the text corpus to pre-train a PLM from scratch, which brings a huge computational overhead, while our PICSO overcomes this weakness and supports flexible continual learning. 
SAPBERT is evaluated with biomedical entity linking solely with the synsets from UMLS, while PICSO is evaluated by four different tasks with synonym knowledge of both general domain and the medical domain.
}

\subsection{Similarity-oriented Tasks}
Similarity-oriented tasks refer to tasks whose core challenge is to capture entity-level or sentence-level semantic similarity, which covers an extensive range of natural language processing tasks. 
Modeling entity-level and sentence-level semantic similarity play a significant role for almost all AI applications, such as machine translation\citep{wieting2019beyond}, dialogue systems \citep{ma2020survey}, and recommendation systems \citep{riyahi2020providing}.
As shown in Table \ref{tab:rw_entity}, entity-level similarity-oriented tasks can be roughly categorized according to the source of two entities (source and target entity) in the entity pairs: (1) Both source and target entities are derived from unstructured text. Typical tasks include entity resolution \citep{peeters2021dual}, lexical simplification \citep{qiang2020lexical}. For example, the task set for entity resolution is to determine whether both entities are identical given two entities and their contextual descriptions (e.g. text, table). (2) Source and target entities are derived from unstructured text and structured KGs, respectively. Representative tasks include entity linking \citep{zhang2010entity}, coreference resolution \citep{lee2017end}, etc. For instance, the goal of entity linking is to link entities in sentences to the corresponding entity entries in KG. (3) Both source and target entities come from structured knowledge graphs, and representative tasks consist of KG canonicalization \citep{dash2021open}, entity alignment \citep{tang2021bert}, and ontology alignment \citep{he2022bertmap}. Entity alignment associates entities in different knowledge graphs if they are semantically same. The core challenge of all the above tasks is to 
model semantic similarity based on entities as well as their contexts. Sentence-level similarity-oriented tasks are likewise an essential branch of general NLU tasks, including paraphrase identification \citep{yin2015convolutional}, semantic textual similarity \citep{cer2017semeval}, and numerous others. For example, semantic textual similarity measures the meaning similarity of sentences, which is also included in the General Language Understanding Evaluation (GLUE) benchmark \citep{wang2018glue}.

\begin{figure*}[ht]
\centering
\includegraphics[width=0.9\textwidth]{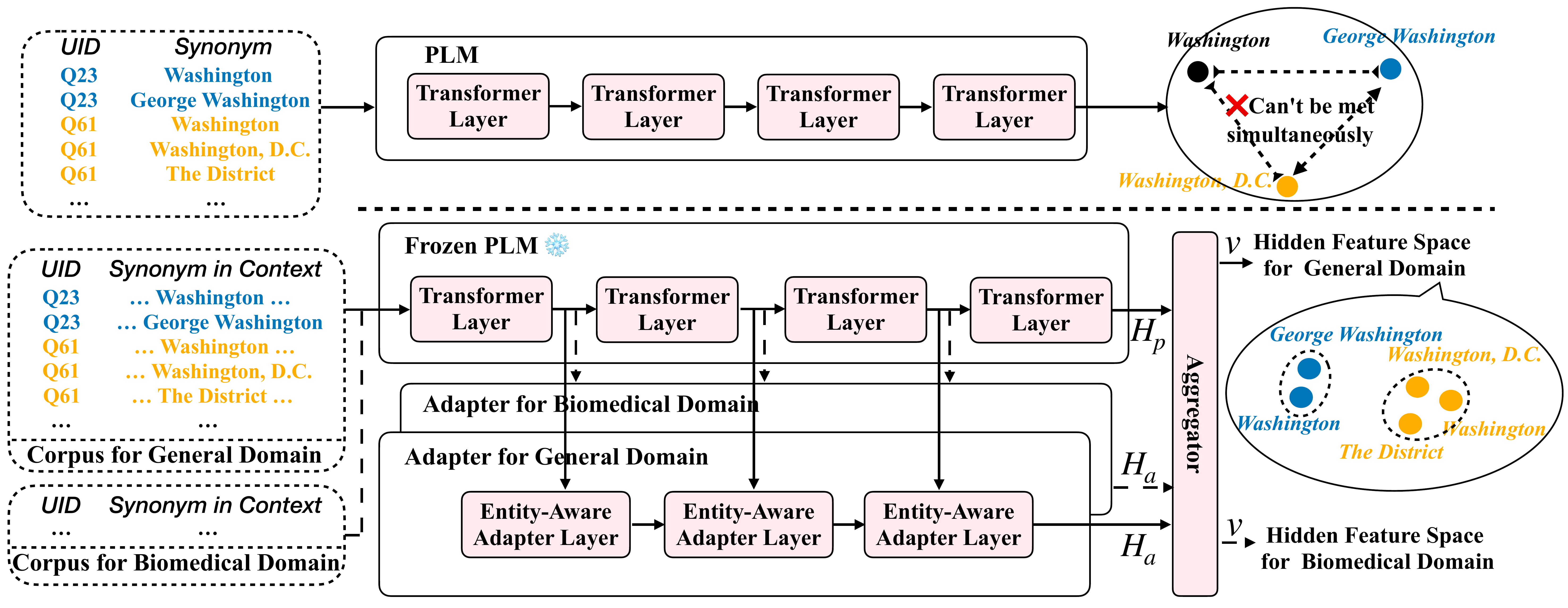}
\caption{\cjy{Frameworks of injecting synonym knowledge into PLMs. Top: the previous methods LIBERT and SAPBERT. Bottom: our method PICSO.}
}
\label{fig:method}
\end{figure*}

\section{\cjy{Methodology}}
\subsection{Overall Framework}
As shown in \cjy{the top of Figure \ref{fig:method}, the framework of the previous methods LIBERT and SAPBERT take the context-missing synonyms as input and directly 
update the parameters of the PLM in pre-training. 
Such a framework cannot handle ambiguous or context-sensitive synonyms, and will inevitably disrupt the original semantic understanding capabilities of the PLM learned from the text corpus.}
To address these issues, \cjy{as shown in the bottom of Figure \ref{fig:method},
our method PICSO uses sentences with marked synonyms rather than individual synonyms as input. In pre-training,}
the internal parameters of the PLM are frozen, while some entity-aware Adapters are attached with each one trained by synonym knowledge of one domain.
We will next introduce some key data annotations and the framework modules.

\cjy{
We assume a collection of synsets $\{S_1, ..., S_m\}$ are extracted from synonym knowledge sources.}
Each synset \cjy{is a collection of synonyms (i.e., words or phrases that have the same meaning), denoted as} $S_i=\{{\rm{UID}}_{i}, e_{i}^{1}, ..., e_{i}^{n}\}$, where ${\rm{UID}}_{i}$ denotes the unique identifier\footnote{\cjy{UID is determined by the synonym knowledge source. For example, in Wikidata, a UID is by the letter Q and a number (e.g., Q61); 
in UMLS, UID is 
by the letter C and a number}.} \cjy{of the synset $S_i$.
}
\cjy{The pre-training corpus $C$ is composed of instances, and each instance is denoted as} $x=\{{\rm{UID}},w, p_s,p_e\}$, where $w$ is a \cjy{sequence of tokens with a marked synonym. 
We} define two special markers $\langle e \rangle$ and $\langle /e \rangle$  to locate the synonym, \cjy{and use $p_s$ and $p_e$ to represent} the indexes of $\langle e \rangle$ and $\langle /e \rangle$ in the sequence, respectively. An instance example is \cjy{as follows}: $x=\{{\rm{Q61}},[...,\langle e \rangle, \mathit{Washington}, \mathit{DC},\langle /e \rangle, ...], 26, 29\}$.

\cjy{PICSO mainly includes three modules. The first module is a \textbf{Frozen PLM}}. We select $\rm{BERT}_{base}$ as the backbone. 
\cjy{It outputs hidden features denoted as $\bm{H}_{p} \in \mathbb{R}^{l\times d}$, in which $l$ and $d$ represent the length of the input word sequence} and the dimension of the last hidden features of BERT, respectively. 
\cjy{The second module is \textbf{Entity-aware Adapter}.
The} input is the hidden features output by Transformer layers of the BERT. 
Each Adapter is plugged into the PLM as a separate module and pre-trained independently for learning \cjy{synonym knowledge of a specific domain.
It learns the semantics of the entities with their contexts and the entire sentence semantics via a combination of two masked self-attention mechanisms, and eventually outputs features $\bm{H}_{a} \in \mathbb{R}^{l\times d}$.}
\cjy{The third module is an \textbf{Aggregator}.
It fuses $\bm{H}_{p}$ and $\bm{H}_{a}$ to obtain the final feature $v$ where two different strategies are proposed.
The modules are pre-trained with a contrastive learning objective, using the pre-training corpus.
It is worth mentioning that the Adapters can be continuously learned with one domain by another, and can be either used together or independently in Aggregator for a downstream task.
}
\subsection{Entity-Aware Adapter}
\begin{figure}[t]
\centering
\includegraphics[width=0.4\textwidth]{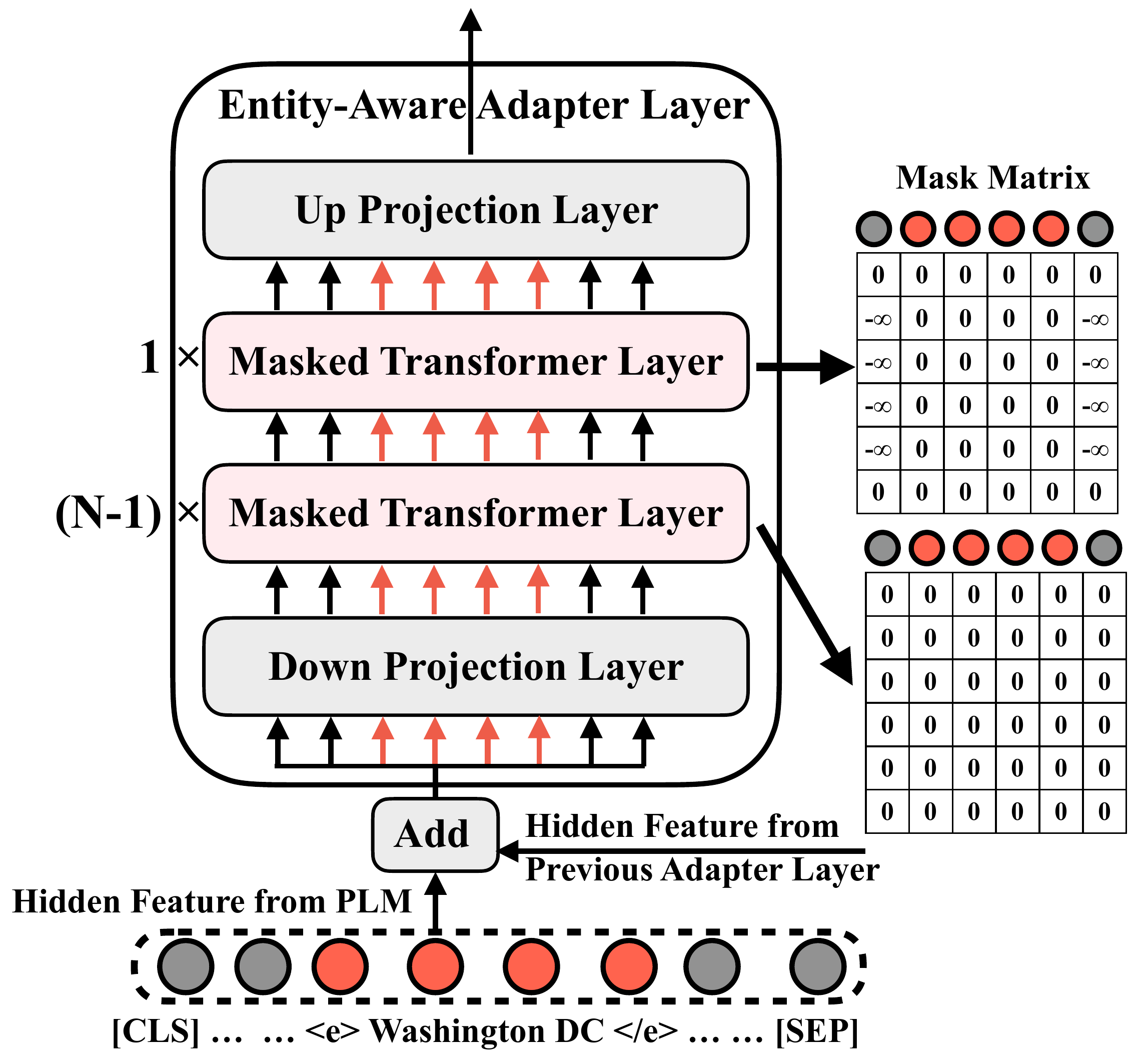}
\caption{\cjy{Structure of one layer of our} entity-aware Adapter.}
\label{fig:adapter}
\end{figure}
\cjy{
Our Adapters are expect to capture the entity-centric semantics from multiple different domains.
Each Adapter contains $K$ layers, while each layer is a stack of one down projection layer, $N$ Transformer layers and one up projection layer, as shown in Figure \ref{fig:adapter}.}
The output of the intermediate Transformer layer of the frozen PLM and the output of the previous Adapter layer are summed up as the input of the current Adapter layer, \cjy{where a residual connection is applied between the input and the output. 
To enable} the Adapter to perceive the semantics of entities  \cjy{with their contexts, a new extension of the} masked self-attention mechanism is proposed:
\begin{equation}
\begin{gathered}
q^{l+1}_{i}, k^{l+1}_{i}, v^{l+1}_{i}=h^{l}_{i} W_{q}, h^{l}_{i} W_{k}, h^{l}_{i} W_{v} \\
h^{l+1}_{i} = \operatorname{Attention}\left( q^{l+1}_{i}, K^{l+1}, V^{l+1} \right)\\
=\sum_{j=1}^{l} \frac{1}{Z} \exp \left(\frac{\left\langle q^{l+1}_{i}, k^{l+1}_{j}\right\rangle + M^{l+1}_{ij}}{\sqrt{d_{k}}}\right) v^{l+1}_{j} \\
\end{gathered}
\end{equation}
where $h_{i}^{l} \in \mathbb{R}^{d}$ represents the hidden feature corresponding to the $i$-th token of the $l$-th Transformer layer. $W_{q}$, $W_{k}$, $W_{v}$ are trainable parameters. $Z$ and $\sqrt{d_{k}}$ refer to the normalization factor and scale factor, respectively. $M^{l}$ is the mask matrix of the $l$-th Transformer layer. 
\cjy{Note that when $M^{l}_{ij}$ tends to $-\infty$, the value of $\exp \left( * \right)$ tends to 0, that is, the token $w_i$ is not concerned with the semantics of the token $w_j$.
When} $M^{l}_{ij} = 0$, the computation degenerates to a regular self-attention mechanism. Formally, the mask matrix $M$ is defined as:
\begin{equation}
M^{l}_{ij}= \begin{cases} -\infty & l = N \land p_{s} \leq i \leq p_{e} \land j \textless p_{s} \\ -\infty & l = N \land p_{s} \leq i \leq p_{e} \land j \textgreater p_{e} \\ 0 & Otherwise \end{cases}
\end{equation}
This means that for the first $N$-1 Transformer layers, we emmploy the conventional self-attention mechanism such that the hidden features model the semantic features of the whole sentence. For the last Transformer layer, \cjy{the new} masked self-attention mechanism is adopted to make each token of entity between $\langle e \rangle$ and $\langle /e \rangle$ focus on its own entity sense, achieving a trade-off between sentence sense and entity sense.

\subsection{Aggregator} 
The output features of the frozen PLM and the Adapter are fed to the aggregator to obtain the final features $v$. During pre-training, the aggregator concatenates $\bm{H}_{p}$ and $\bm{H}_{a}$ and takes out the vectors at indexes $p_s$ and $p_e$ to be concatenated again:
$\bm{H}=\bm{H}_{p} \oplus \bm{H}_{a}, h=\bm{H}[p_s] \oplus \bm{H}[p_e] $. Then, the resulting intermediate feature ${h} \in \mathbb{R}^{4d}$ is passed into a fully-connected layer followed by a normalization layer. \cjy{When applied in downstream tasks}, if fine-tuning is performed, we send the concatenated features $\bm{H}$ into the task-specific layer. The Adapter can be used individually. If multiple Adapters are used, $\bm{H}_{p}$ and multiple $\bm{H}_{a}$ can be concatenated together. \cjy{If PICSO is directly used} as a feature extractor without fine-tuning, we sum $\bm{H}_{p}$ and the $l_2$-normalized $\bm{H}_{a}$ as the semantic features for the downstream task.

\subsection{Pre-training Objective}
\cjy{We apply a contrastive objective which pulls synonymous pairs closer and non-synonymous pairs away to train the Adapters and the aggregator.}
To do this we first \cjy{generate} positive and negative instance pairs from \cjy{each batch of the pre-training corpus. For an arbitrary instance $x_i$ in the batch, it is combined with each of the instances that have the same UID as $x_i$ (i.e., $pos(x_i) = \left\{x_j \mid x_{i}[UID]=x_{j}[UID]\right\}$) for positive instance pairs,
and combined with the other instances in the batch for negative instance pairs.
}
Inspired by \citep{robinson2020hard,li2022past,10.1145/3477495.3531954}, we design the contrastive objective, which concerns more on the hard negative instance pairs, i.e., non-synonymous pairs that are difficult to distinguish.
Concretely, the contrast loss is calculated as follows:
\begin{equation}
\begin{gathered}
\mathcal{L}_{c l}=-\sum_{i=1}^{B} \log \frac{S_{i}^{+}}{S_{i}^{+}+S_{i}^{-}}, \\
S_{i}^{+}=\sum_{j=1}^{\left | pos(x_{i}) \right | }e^{\mathrm{v}_{i}^{\top} \cdot \mathrm{v}_{j} / t}, \\
S_{i}^{-}=\max \left(\frac{-(B-1-\left | pos(x_{i}) \right |) \cdot \tau^{+} \cdot S_{i}^{+}+\widetilde{S_{i}^{-}}}{1-\tau^{+}}, e^{\frac{-1}{t}}\right), \\
\widetilde{S_{i}^{-}}=\frac{(B-1-\left | pos(x_{i}) \right |) \sum_{k: k \neq i \neq pos(x_{i})} e^{(1+\beta) \mathrm{v}_{i}^{\top} \cdot \mathrm{v}_{k} / t}}{\sum_{k: k \neq i \neq pos(x_{i})} e^{\beta \mathrm{v}_{i}^{\top} \cdot \mathrm{v}_{k} / t}}.
\end{gathered}
\end{equation}
where ${S_{i}^{+}}$ \cjy{(resp. ${S_{i}^{-}}$)} reflects the similarity between training pairs from the positive \cjy{(resp. negative)} pairs, $B$ is the size of \cjy{a batch},  $\tau^{+}$ is the class-prior probability \cjy{that} can be estimated from data or treated as a
hyper-parameter, $\beta$ is the hyper-parameter controlling the level of concentration on negative samples, $t$ is the temperature scaling factor which we set as $0.5$ in all our experiments. 
The $\widetilde{S_{i}^{-}}$ term awards higher weights to negative \cjy{instance pairs whose instances have high similarity (i.e., hard negative samples) by reweighting.
We assign} a greater penalty to these hard negative samples instead of mining hard negative samples by modifying the sampling strategy as in LIBERT and SAPBERT. Hence, our proposed contrastive objective is easier to use and achieves better results, as also demonstrated in Section \ref{sec:loss}.

\subsection{Construct Pre-training Corpus}
\cjy{One entity-aware Adapter learns  independently from the corpus of one domain. We consider a general domain with the corpus extracted from Wikidata and a biomedical domain with the corpus extracted from UMLS.
The concrete corpus construction procedure has the following} two steps:

\noindent\textbf{Gathering sentences \cjy{and} synonyms.} 
\cjy{We get the synsets from the two knowledge bases, Wikidata and UMLS, which already have a massive collection of high-quality synonyms (i.e., entities with identical meanings). 
To construct the corpus, we further collect sentences that contain entity mentions linked to knowledge base entities.}
For the general domain, \cjy{we crawl Wikipedia articles that}
contain abundant \cjy{entity mentions with human-annotated hyperlinks to Wikidata entities.}
For the biomedical domain, we use the off-the-shelf high-precision entity linking tool Medlinker \citep{loureiro2020medlinker} to link entity mentions of article abstracts from PubMed\footnote{https://pubmed.ncbi.nlm.nih.gov/} to UMLS. In addition, we include the definitions of entities from UMLS as a supplementary corpus. We also remove simple pairs with edit distances less than 10 and limit the generation of up to 50 synonym pairs per UID.

\noindent\textbf{Balancing low-frequency entities.} To ensure the proportion of low-frequency synonyms in the corpus, we replaced entities in some sentences with \cjy{their} low-frequency synonyms. For example, \textit{California} in \cjy{a sentence would be replaced by} \textit{The Golden State}.

\cjy{In the end, we constructed over 3.7 million and over 12.8 million context-equipped synonym pairs for the general domain and the biomedical domain, respectively.}
The statistics of the pre-training corpus for each domain are shown in Table \ref{tab:data_pretrain}.

\begin{table}[ht]
\centering
\caption{Statistics for pre-training corpus}
\scalebox{0.6}{
\begin{tabular}{lccccc}
\toprule
\textbf{Domain} & \textbf{\# UID} & \textbf{\# Synonym Pairs} & \textbf{\shortstack{\# Synonym \\ Pairs per UID}} & \textbf{\shortstack{Average \\ Sentence Len}} & \textbf{\shortstack{Avgerage \\Edit Distance}} \\ \midrule
General         & 0.65M           & 3.75M                     & 5.8                            & 117.9            & 18.4              \\
Biomedical      & 2.10M           & 12.79M                    & 6.1                            & 107.3            & 27.7              \\ \bottomrule 
\end{tabular}}
\label{tab:data_pretrain}
\end{table}

\section{Experiments}
\label{sec:exp}
\subsection{Experiment Setup}
\subsubsection{Downstream Tasks.} 
We chose four extensively studied similarity-oriented \cjy{downstream tasks, i.e.,} entity resolution, entity linking, KG canonicalization, and lexical simplification, covering the assessment of \cjy{tasks consuming} both structured data and unstructured text. 
\cjy{For entity resolution and entity linking, PICSO is further fine-tuned by their samples; while for the other two tasks, PICSO is not fine-tuned, which evaluates PICSO in a zero-shot or unsupervised setting.}
To verify that \cjy{PICSO also benefits general} NLU tasks, we also conducted experiments on GLUE which has 9 NLU tasks.
Each task will be separately introduced in detail bellow.

\subsubsection{\cjy{Baselines.}} 
We comprehensively compare \cjy{PICSO with three types of baselines}. \textbf{Basic PLMs}: (1) \textbf{BERT} \citep{kenton2019bert} is an important baseline since PICSO is based on it; (2) \textbf{Roberta} \citep{liu2019roberta} is the the advanced version of BERT, which removes the next sentence prediction task and employs a larger pre-trained corpus. \textbf{\cjy{PLMs with KG knowledge injected}}: (3) \textbf{ERNIE-THU} \citep{zhang2019ernie} is the most \cjy{classic} KG knowledge-enhanced PLM model, which incorporates entity representations learned through TransE \cjy{into BERT;} (4) \textbf{K-Adapter} \citep{wang2021k} \cjy{injects KG structured knowledge by an Adapter via relational classification.} 
Note that our entity-aware Adapter is an enhancement of the original Adapter. \cjy{The original K-Adapter is} based on $\rm{Roberta}_{large}$. For a fair comparison, we re-implemented K-Adapter with $\rm{Roberta}_{base}$. \textbf{PLMs \cjy{with synonym knowledge injected}}: (5) \textbf{LIBERT} \citep{lauscher2020specializing} train\cjy{s} BERT from scratch \cjy{by predicting synonymous entity pairs;
(6) \textbf{SAPBERT} \citep{liu2021self} continues the pre-training of a BERT with massive synonymous and non-synonymous entity pairs from UMLS. 
Since LIBERT and SAPBERT are the most relevant methods to this study, they will get more attention in the following result analysis. Note that PICSO is compatible for the methods designed for specific tasks, so we don't compare task-specific SOTA methods, as in much of the PLM works.
}

\subsubsection{PICSO Setup}
We use $\rm{BERT}_{base}$ 
which contains 12 Transformer layers and 433M parameters. The dimension of the hidden feature corresponding to each token is 768. Each entity-aware Adapter contains 3 Adapter layers plugged at layers \cjy{0, 5, 11 of BERT}. Two Transformer layers identical to those in BERT are \cjy{set in} each Adapter layer, i.e., $N$=2. We tally the parameters for each entity-aware Adapter to be approximately 46M.
Compared \cjy{with} LIBERT and SAPBERT, which require pre-training the entire BERT, we have \cjy{fewer parameters to tune}. 
We pre-train the PICSO for 3 epochs on 8 Tesla V100s with a batch size of 256. The time to train \cjy{the general domain Adapter and the medical domain Adapter is about 0.46 h and 1.58 h per epoch}, respectively. The full model \cjy{using} two Adapters is abbreviated as PICSO(W+U). The model \cjy{using one Adapter trained on Wikidata (resp. UMLS) is  denoted as PICSO(W) (resp. PICSO(U)).} PICSO(w/o k) represents BERT with a randomly initialized Adapter, i.e., an Adapter without synonym knowledge injection.

\subsection{Experiments on Similarity-oriented Tasks}
\subsubsection{Entity Linking}
\paragraph{Datasets and Fine-tuning.} Entity linking aims to \cjy{match an entity mention in a textual context with an entity in the target KG.}
SAPBERT \cjy{is} also tested on the entity linking task, but the datasets used such as BC5CDR, \cjy{lack the entity mentions' contexts} and are less ambiguous. 
Hence, we \cjy{adopt a} more challenging and widely used \cjy{dataset named} AIDA CoNLL-YAGO \citep{hoffart2011robust},
\cjy{and conduct} cross-domain experiments following the same setting \cjy{as} in \citep{le2018improving,yang2019learning,zhang2022hsie}. \cjy{Namely,} models are trained on \cjy{a training subset of AIDA (AIDA-train) and evaluated on a test subset of AIDA as well as} five popular public datasets: AQUAINT, MSNBC, ACE2004, CWEB and WIKI, which cover a wide range of domains such as medicine \cjy{and technology} \& science. 
To fine-tune the PLMs for entity linking, as with SAPBERT, we use the multi-similarity loss, a metric learning objective that adjusts the pairwise distances of positive and negative pairs. Each positive pair consists of \cjy{an entity mention and its corresponding entity} in KG, and negative pairs are generated by randomly corrupting positive pairs. Acc@k is employed to evaluate the \cjy{a model with the ranking of KG entities,} which means the percentage of the top-k predictions that contain the ground truth KG entity. 
To ensure fairness, for all downstream tasks, we set their key hyperparameters (e.g., learning rate, number of training rounds), the same, and no complex tuning of hyperparameters is performed.

\paragraph{Results.} \cjy{The results for entity linking are shown in Table \ref{tab:linking}, through which we can have the following observations.} 
(1) PICSO(W+U) achieves the best results on all six datasets, and the \cjy{absolute} improvement of Acc@1 on some datasets even exceeds \cjy{5\%}. 
Except for LIBERT on AQUAINT,  LIBERT and SAPBERT \cjy{show general performance decreasements}, 
which confirms \cjy{that their pre-training negatively impacts the original semantic understanding  capabilities of PLMs}. (2) \cjy{Injecting KG knowledge does not necessarily benefits}
entity linking. K-Adapter outperforms its \cjy{base PLM} Roberta in some cross-domain cases, while ERNIE-THU generally shows a light degradation compared to its \cjy{base PLM} BERT. 
(3) PICSO (W) has a higher gain on performance than PICSO (U).
Although these \cjy{testing datasets contain medical data, medical data only take a small ratio, and the general synonym knowledge from Wikidata contributes more}. 
(4) \cjy{PICSO (w/o k) \todo{has an overall gain compared with the base PLM BERT}. This is due to the fine-tuning and a larger model of PICSO.}

\begin{table*}[ht]
\centering
\caption{Results \cjy{(\%)} on entity linking
}
\scalebox{0.9}{
\begin{tabular}{l|cc|cc|cc|cc|cc|cc}
\toprule
\multirow{2}{*}{\textbf{Model}} & \multicolumn{2}{c|}{\textbf{AIDA}} & \multicolumn{2}{c|}{\textbf{ACE2004}} & \multicolumn{2}{c|}{\textbf{AQUAINT}} & \multicolumn{2}{c|}{\textbf{CWEB}} & \multicolumn{2}{c|}{\textbf{MSNBC}} & \multicolumn{2}{c}{\textbf{WIKI}} \\ \cline{2-13} 
                       & \textbf{Acc@1}       & \textbf{Acc@5}      & \textbf{Acc@1}       & \textbf{Acc@5}        & \textbf{Acc@1}       & \textbf{Acc@5}       & \textbf{Acc@1}       & \textbf{Acc@5}     & \textbf{Acc@1}       & \textbf{Acc@5}       & \textbf{Acc@1}       & \textbf{Acc@5}      \\ \midrule
BERT                   & 72.35       & 86.64      & 70.81        & 82.87        & 60.24        & 80.74        & 47.84       & 68.59      & 56.40       & 78.04       & 58.68       & 79.15      \\ \midrule
Roberta                & 74.89       & 87.32      & 67.91        & 79.64        & 58.28        & 78.93        & 45.22       & 65.54      & 55.25       & 75.54       & 55.27       & 76.13      \\ \midrule
ERNIE-THU              & 69.12       & 86.48      & 69.26        & 80.93        & 51.71        & 73.03        & 44.49       & 65.39      & 56.25       & 77.28       & 55.97       & 76.98      \\ \midrule
K-Adapter              & 73.72 & 85.72 & 69.84 & 80.41 & 60.27 & 79.86 & 45.49 & 65.37 & 53.11 & 72.93 & 53.56 & 75.70        \\ \midrule
LIBERT                 & 69.69       & 83.75      & 66.92        & 78.21        & 63.68        & 80.05        & 48.51       & 66.71      & 53.35       & 74.84       & 57.70       & 76.78      \\ \midrule
SAPBERT                & 54.21       & 73.64      & 49.02        & 70.03        & 52.26        & 77.02        & 37.97       & 60.99      & 41.31       & 66.46       & 47.15       & 66.68      \\ \midrule\midrule
PICSO(W+U)         & \textbf{78.26}       & \textbf{88.63}      & \textbf{75.87}        & \textbf{84.26}        & \textbf{69.32}        & \textbf{82.46}        & \textbf{52.25}       & \textbf{70.17}      & \textbf{62.50}       & \textbf{81.47}       & \textbf{61.04}       & \textbf{80.69}      \\ \midrule
PICSO(W)           & 76.79       & 88.25      & 73.64        & 83.87        & 66.71        & 81.70        & 49.49       & 69.52      & 60.04       & 80.62       & 60.33       & 80.28      \\ \midrule
PICSO(U)           & 75.91       & 88.05      & 72.84        & 83.65        & 63.54        & 81.15        & 48.62       & 69.39      & 59.06       & 79.47       & 58.93       & 79.34  \\  \midrule
PICSO(w/o k)           & 73.10        & 87.10        &71.98         &83.20          & 63.13          & 80.90          & 48.35        & 68.21        & 52.28       & 72.56       & 57.15        & 76.60    \\ \bottomrule
\end{tabular}}
\label{tab:linking}
\end{table*}

\subsubsection{Entity Resolution}
\paragraph{Datasets and Fine-tuning.} Entity resolution 
(a.k.a entity matching) 
is \cjy{to find} records that refer to the same real-world entity across different data sources.
Following \citep{peeters2021dual}, we compare \cjy{the methods} 
using three datasets: WDC LSPC, \cjy{DBLP-Scholar and Company}. The WDC LSPC \cjy{dataset is} built by product offers from e-shop, containing \cjy{four categories of products: computer, camera, shoe} and watch. 
\cjy{For each product category, there are one test set and four training sets with different sizes.}
In other words WDC LSPC has 4$\times$4 sub-datasets. We pick two training sets of medium (M) and large (L) sizes for each category in our experiments.
For \cjy{WDC LSPC}, the entities in the test set all appear in the training set, while for \cjy{Company and DBLP-Scholar, there is no overlap between the entities of the test set and those of the training sets.
Thus, WDC LSP enables the evaluation with \textit{seen} entities, while the other two enable the evaluation with \textit{unseen} entities.}
\cjy{Following \citep{peeters2021dual}, a
linear layer and a sigmod function are attached after each pre-trained model (on top of the hidden feature of the [CLS] token) to predict whether two entities match or not, and a binary cross-entropy loss is used to fine-tune the model.
Precision (P), Recall (R) and F1 Score are adopted as the metrics.}

\paragraph{Results.} The results for entity resolution are shown in Table \ref{tab:matching}, \cjy{from which we can have similar observations as in entity linking.}
We note that in the seen case, i.e., in the WDC LSPC, PICSO is insensitive to the size of the training set compared to other PLMs, suggesting that the synonyms contain \cjy{partial} knowledge that entity resolution desires. The testing set sizes of Company and DBLP-Scholar are 20 and 5 times larger than \cjy{that of} WDC LSPC, respectively. Thus the improvement by PICSO in the unseen case is actually more significant compared to the seen case, demonstrating the sound learning potential of PICSO. Compared to PICSO, LIBERT and SAPBERT have competitive recall \cjy{but much worse precision. This} 
indicates that LIBRT and SAPBERT confusingly treat some non-synonymous pairs as synonymous, which \cjy{could be} caused by the lack of contextual constraints in pre-training.

\begin{table*}[ht]
\centering
\caption{Results \cjy{(\%) on entity resolution}
}
\scalebox{0.55}{
\renewcommand\arraystretch{1}
\setlength\tabcolsep{2.5pt}
\begin{tabular}{l|ccc|ccc|ccc|ccc|ccc|ccc|ccc|ccc|ccc|ccc}
\toprule
\multirow{2}{*}{\textbf{Model}} & \multicolumn{3}{c|}{\textbf{Computers-M}} & \multicolumn{3}{c|}{\textbf{Computers-L}} & \multicolumn{3}{c|}{\textbf{Shoes-M}} & \multicolumn{3}{c|}{\textbf{Shoes-L}} & \multicolumn{3}{c|}{\textbf{Watches-M}} & \multicolumn{3}{c|}{\textbf{Watches-L}} & \multicolumn{3}{c|}{\textbf{Cameras-M}} & \multicolumn{3}{c|}{\textbf{Cameras-L}} & \multicolumn{3}{c|}{\textbf{Company}} & \multicolumn{3}{c}{\textbf{DBLP-Scholar}} \\ \cline{2-31} 
                       & \textbf{P}     & \textbf{R}     & \textbf{F1}       & \textbf{P}     & \textbf{R}     & \textbf{F1}       & \textbf{P}     & \textbf{R}     & \textbf{F1}      & \textbf{P}     & \textbf{R}     & \textbf{F1}      & \textbf{P}     & \textbf{R}     & \textbf{F1}      & \textbf{P}     & \textbf{R}     & \textbf{F1}      & \textbf{P}     & \textbf{R}     & \textbf{F1}      & \textbf{P}     & \textbf{R}     & \textbf{F1}      & \textbf{P}     & \textbf{R}     & \textbf{F1}      & \textbf{P}     & \textbf{R}     & \textbf{F1}       \\ \midrule
BERT                   & -         & -        & 89.31    & -         & -        & 92.11    & -       & -       & 79.82   & -       & -       & 87.37   & -        & -        & 89      & -        & -        & 95.23   & -        & -        & 87.02   & -        & -        & 91.02   & -       & -       & 91.7    & -         & -         & 95.27    \\ \midrule
RoBERTa                & -         & -        & 91.9     & -         & -        & 94.68    & -       & -       & 81.12   & -       & -       & 86.6    & -        & -        & 92.28   & -        & -        & 93.93   & -        & -        & 90.2    & -        & -        & 93.91   & -       & -       & 91.81   & -         & -         & 95.29    \\ \midrule
ERNIE-THU              & 80.00        & 93.33    & 86.15    & 90.9      & 93.33    & 92.1     & 78.41   & 82.6    & 80.45   & 90.23   & 87.33   & 88.75   & 77.59    & 92.33    & 84.32   & 93.53    & 91.66    & 92.59   & 76.98    & 90.33    & 83.12   & 87.61    & 92.00       & 89.75   & 89.29   & 93.81   & 91.5    & 95.14     & 95.32     & 95.23    \\ \midrule
K-Adapter             & 91.55 & 91.55 & 91.55 & 94.95 & 94.33 & 94.63 & 87.45 & 86.28 & 86.86 & 95.10 & 90.96 & 92.99 & 90.28 & 96.00 & 93.05 & 97.50 & 91.33 & 94.32 & 92.00 & 92.00 & 92.00 & 93.85 & 91.66 & 92.74 &93.27 &92.09 &92.67 & 94.69 & 96.82 & 95.74            \\ \midrule
LIBERT                 & 79.41     & 90.00       & 84.37    & 86.29     & 92.33    & 89.21    & 77.23   & 83.94   & 80.44   & 86.97   & 82.60    & 84.73   & 84.37    & 90.00       & 87.09   & 94.82    & 91.66    & 93.22   & 79.68    & 83.66    & 81.62   & 87.91    & 87.33    & 87.62   & 76.9    & 92.76   & 84.09   & 93.21     & 94.95     & 94.07    \\ \midrule
SAPBERT                & 53.61     & 81.66    & 64.72    & 52.60      & 87.66    & 65.75    & 76.87   & 82.27   & 79.48   & 87.01   & 82.94   & 84.93   & 87.37    & 90.00       & 88.66   & 94.36    & 89.33    & 91.78   & 87.26    & 91.33    & 89.25   & 86.16    & 91.33    & 88.67   & 90.74   & 89.89   & 90.31   & 94.10      & 96.91     & 95.48    \\ \midrule \midrule
PICSO(W+U)         & 93.62 & 93.00 & \textbf{93.31} & 94.50 & 95.66 & \textbf{95.07} & 91.17 & 92.62 & \textbf{92.62} & 94.88 & 92.97 & \textbf{93.91} & 93.77 & 94.00 & \textbf{93.88} & 98.57 & 95.00 & \textbf{96.75} & 94.68 & 93.42 & \textbf{94.04} & 95.84 & 92.33 & \textbf{94.05} & 92.94 & 92.91 & \textbf{92.92} & 94.26 & 97.87 & \textbf{96.03}    \\ \midrule
PICSO(W)           & 90.70      & 94.33    & 92.48    & 94.96     & 94.33    & 94.64    & 91.97   & 91.97   & 91.97   & 95.12   & 91.3    & 93.17   & 93.64    & 93.33    & 93.48   & 97.93    & 94.66    & 96.27   & 93.19    & 91.33    & 92.25   & 94.33    & 94.33    & 94.33   & 92.63   & 92.3    & 92.46   & 94.32     & 97.56     & 95.91    \\ \midrule
PICSO(U)           & 92.35     & 92.66    & 92.51    & 93.66     & 93.66    & 93.66    & 91.08   & 92.3    & 91.69   & 94.03   & 89.63   & 91.78   & 92.66    & 92.66    & 92.66   & 98.58    & 93.00       & 95.71   & 93.11    & 94.66    & 93.88   & 94.86    & 92.33    & 93.58   & 92.87   & 92.42   & 92.65   & 94.09     & 97.34     & 95.69    \\ \midrule 
PICSO(w/o k)      & 91.14 & 92.66 & 91.90 & 91.65 & 94.03 & 92.82 & 84.51 & 87.62 & 86.04 & 93.07 & 89.30 & 91.00 & 91.64 & 92.33 & 91.98 & 96.91 & 96.91 & 95.60 & 91.88 & 89.66 & 90.75 & 94.16 & 90.00 & 92.03 & 91.02 & 93.61 &92.29 & 94.49 & 95.74 & 95.11     \\ \bottomrule
\end{tabular}}
\label{tab:matching}
\end{table*}

\subsubsection{Lexical Simplification}
\paragraph{Datasets and \cjy{Task Method}.} Lexical simplification seeks to replace target words in contextual sentences with simpler substitutes without altering the \cjy{meanings. It is an important evaluation task in LIBERT.} 
Three public datasets  \cjy{--- LexMTurk, BenchLS, and NNSeval are used for evaluation} 
where LexMTurk is collected from Wikipedia, and BenchLS and NNSeval are expanded versions of LexMTurk. 
\cjy{As the evaluation of LIBERT, an unsupervised pre-trained PLM-based method \citep{qiang2020lexical}, which
first generates a set of candidate substitutes and then ranks these substitutes, is adopted.
We report scores of Precision, Recall and F1 score for candidate generation, as well as the scores of Accuracy for the final result, so as to compare different pre-trained PLMs that are adopted.
} 

\paragraph{Results.} The results \cjy{of} lexical simplification are presented in Table \ref{tab:simply}.
\cjy{ We can observe that PICSO obtains the highest final accuracy on all three datasets.}
Surprisingly, SAPBERT scores almost 0 on all three datasets. On the one hand, SAPBERT severely loses its semantic understanding capability of textual context, and on the other hand, the unsupervised \cjy{setting} prevents it from reconstructing the semantic space by fine-tuning. 
\cjy{Regarding LIBERT, it has a lead in precision for substitute generation on BenchLS and LexMTurk}. However, this is achieved at the expense of recall, \cjy{which is undesirable in the first stage of substitute generation.}
\cjy{Meanwhile,} although LIBERT \cjy{authors claim} that it outperforms their own implementation of BERT with a smaller pre-training corpus, it has no advantage over the original BERT implemented by Google.
%

\begin{table*}[ht]
\centering
\caption{Results \cjy{(\%) on lexical simplification}
}
\scalebox{1}{
\begin{tabular}{l|cccc|cccc|cccc}
\toprule
\multirow{2}{*}{Model} & \multicolumn{4}{c|}{\textbf{BenchLS}}   & \multicolumn{4}{c|}{\textbf{LexMTurk}}  & \multicolumn{4}{c}{\textbf{NNSeval}}   \\ \cmidrule{2-13} 
                       & \textbf{P}     & \textbf{R}     & \textbf{F1}    & \textbf{Acc}   & \textbf{P}     & \textbf{R}     & \textbf{F1}    & \textbf{Acc}   & \textbf{P}     & \textbf{R}     & \textbf{F1}    & \textbf{Acc}   \\ \midrule
BERT                   & 24.81 & 32.08 & 27.98 & 57.03 & 32.04 & 23.43 & 27.06 & 71.24 & 18.32 & 23.78 & 20.69 & 37.65 \\ \midrule
Roberta                & 20.08 & 27.25 & 23.12 & 52.41 & 25.66 & 19.95 & 22.45 & 64.00 & 18.87 & 25.18 & 21.57 & 41.84 \\ \midrule
ERNIE-THU              & 24.19 & 32.83 & 27.86 & 55.97 & 31.60 & 24.57 & 27.64 & 70.00 & 18.66 & 24.90 & 21.33 & 33.05 \\ \midrule
K-Adapter              & 19.93     & 27.05     & 22.95     & 55.40     & 25.28     & 19.66     & 22.11     & 69.32     & 16.23     & 21.66     & 18.56     & 41.09     \\ \midrule
LIBERT                 & \textbf{27.66} & 22.52 & 24.83 & 47.57 & \textbf{37.00} & 17.27 & 23.54 & 63.2  & 15.39 & 20.54 & 17.60 & 27.19 \\ \midrule
SAPBERT                & 0.05  & 0.07  & 0.06  & 0.10  & 0.06  & 0.04  & 0.05  & 0.2   & 0.04  & 0.05  & 0.05  & 0     \\ \midrule \midrule
PICSO(W+U)         & 25.18 & \textbf{34.18} & \textbf{29.00} & \textbf{59.63} & 33.08 & \textbf{25.72} & \textbf{28.94} & 73.20  & \textbf{20.15} & \textbf{26.38} & \textbf{22.84} & \textbf{42.76} \\ \midrule
PICSO(W)           & 25.15 & 34.13 & 28.96 & 58.66 & 32.58 & 25.33 & 28.50 & \textbf{73.59} & 19.03 & 25.40 & 21.76 & 42.65 \\ \midrule
PICSO(U)           & 25.10 & 34.06 & 28.90 & 57.91 & 32.60 & 25.35 & 28.52 & 73.0  & 18.99 & 25.34 & 21.71 & 42.10 \\ \midrule
PICSO(w/o k)       & 20.97       & 28.46           & 24.15      & 47.68       & 28.56      & 22.21       & 24.98 &63.60      &  15.48     & 20.65      & 17.69      & 29.70     \\ \bottomrule
\end{tabular}}
\label{tab:simply}
\end{table*}

\subsubsection{KG Canonicalization}
\paragraph{Datasets and \cjy{Task Method}.} KGs constructed from unstructured data usually store redundant \cjy{entities. KG canonicalization aims to identifying the equivalent entities in a KG.
It is an inherently unsupervised task since we usually are not given any annotated data.} 
As in \citep{dash2021open}, we used four datasets --- Rever-base, Reverb45k, Reverb-ambiguous, and CANONICNELL for evaluation. The first three are homogeneous, while CANONICNELL is \cjy{heterogeneous, constructed based on NELL \citep{carlson2010toward}. A simple PLM-based method \citep{vashishth2018cesi,dash2021open} is often adopted for comparing the performance of PLMs. Briefly it first uses a PLM to build text embeddings for entities} and then uses a Hierarchical Agglomerative Clustering (HAC) algorithm for clustering. We use the macro and micro F1 scores as evaluation metrics.
%
\paragraph{Results.}  
\cjy{The results for KG canonicalization are} shown in Table \ref{tab:cano}. Since KG canonicalization is an intra-KG task and excludes textual context, \cjy{models injected with both synonym knowledge and KG knowledge obtain performance boosts versus their base PLMs.
PICSO achieves the best result, followed by SAPBERT,
} 
which strongly confirms the superiority of synonym knowledge over generic KG knowledge for \cjy{this similarity-oriented task. 
Meanwhile, according to our statistics, medical-related entities in the Reverb* datasets occupy 20\% to 30\%, which is consistent with that PICSO(U) has higher performance than PICSO(W). 
As expected, PICSO(w/o k) shows a serious performance decline} due to feature space corruption.

\begin{table*}[ht]
\centering
\caption{Results \cjy{(\%)} on KG canonicalization
}
\scalebox{1}{
\begin{tabular}{l|cc|cc|cc|cc}
\toprule
\multirow{2}{*}{\textbf{Model}} & \multicolumn{2}{c|}{\textbf{Reverb-base}} & \multicolumn{2}{c|}{\textbf{Reverb45k}} & \multicolumn{2}{c|}{\textbf{Reverb-ambiguous}} & \multicolumn{2}{c}{\textbf{CANONICNELL}} \\ \cmidrule{2-9} 
                       & \textbf{Macro}       & \textbf{Micro}       & \textbf{Macro}       & \textbf{Micro}      & \textbf{Macro}       & \textbf{Micro}          & \textbf{Macro}       & \textbf{Micro}        \\ \midrule
BERT                   & 69.83         & 92.35         & 16.92        & 75.35        & 60.10             & 81.62           & 63.19         & 66.93         \\ \midrule
Roberta                & 75.75         & 92.71         & 26.48        & 76.83        & 61.55            & 87.97           & 70.63         & 76.01         \\ \midrule
ERNIE-THU              & 72.49         & 92.57         & 19.54        & 76.09        & 61.78            & 87.95           & 65.50          & 69.56         \\ \midrule
K-Adapter                       & 75.17               & 91.28               & 28.10              & 76.58              & 59.13                  & 80.83                  & 71.48                & 77.62                \\ \midrule
LIBERT                 & 66.93         & 91.97         & 16.18        & 75.58        & 61.22            & 86.12           & 62.09         & 68.37         \\ \midrule
SAPBERT                & 82.41         & 93.10          & 41.18        & 79.59        & 61.92            & 87.98           & 74.57         & 80.54         \\ \midrule\midrule
PICSO(W+U)        & \textbf{89.21}         & \textbf{93.56}         & \textbf{44.69}        & \textbf{80.67}        & \textbf{62.67}            & \textbf{88.11}           & \textbf{76.39}         & \textbf{81.98}         \\ \midrule
PICSO(W)        & 85.31         & 93.21         & 42.62        & 79.97        & 62.51            & 88.10            & 75.81         & 81.69         \\ \midrule
PICSO(U)        & 87.14         & 93.26         & 43.97        & 79.79        & 62.41            & 88.09           & 76.22         & 81.85         \\ \midrule
PICSO(w/o k)        & 63.10         & 90.84         & 15.70  & 70.74        & 57.54    & 72.44                 & 60.43        & 64.83         \\ \bottomrule
\end{tabular}}
\label{tab:cano}
\end{table*}

\begin{table*}[ht]
\centering
\caption{Results on eight GLUE tasks.}
\scalebox{1}{
\begin{tabular}{l|c|c|c|c|c|c|c|c|c}
\toprule
Model            & \textbf{CoLA}  & \textbf{SST-2} & \textbf{MRPC}  & \textbf{STS-B} & \textbf{QQP}   & \textbf{MNLI-(m/mm)} & \textbf{QNLI}  & \textbf{RTE}   & \textbf{Avg} \\ \midrule
BERT             & 56.53 & 92.32 & 88.85 & 88.48 & 87.49 & 83.81/84.1  & 90.66 & 65.7  & 81.99    \\ \midrule
Roberta     & 50.19 & \textbf{94.15} & 81.83 & 84.88 & 87.48 & \textbf{87.36/87.34} & 92.17 & 56.32 & 80.19    \\ \midrule
ERNIE            & 44.28 & 90.6  & 82.15 & 85.03 & 86.75 & 83.10/83.51 & 89.99 & 58.48 & 78.21     \\ \midrule
K-Adapter        & 54.70 & 93.69 & 85.62 & 87.62 & 86.72 & 87.29/87.02 & \textbf{92.7} & \textbf{68.95} & 82.70     \\ \midrule
SAPBERT          & 4.38  & 88.53 & 81.21 & 82.57 & 85.86 & 81.81/82.54 & 89.38 & 54.87 & 72.35     \\ \midrule
LIBERT           & 37.2  & 89.3  & 88.7  & -     & \textbf{90.0}    & 79.6/80.0   & 87.7  & 66.4  & 77.36     \\ \midrule \midrule
PICSO(W+U)   & 58.04 & 93.89 & 89.41 & \textbf{89.43} & 88.58 & 84.66/84.92 & 91.63 & 64.62 & \textbf{82.80}     \\ \midrule
PICSO(W)     & 57.78 & 92.89 & \textbf{89.83} & 89.12 & 88.07 & 84.61/84.95 & 91.12 & 64.98 & 82.59    \\ \midrule
PICSO(U)    & \textbf{58.29} & 92.32 & 89.61 & 89.09 & 87.82 & 84.63/85.31 & 91.27 & 63.90 & 82.47     \\ \midrule
PICSO(w/o k) & 57.78      & 93.00      &    88.93   & 88.98      & 86.92      & 81.85/82.27      & 90.74       & 63.18  & 81.51  \\ \bottomrule
\end{tabular}}
\label{tab:glue}
\end{table*}
\subsection{Experiments on General NLU Tasks}
The General Language Understanding Evaluation
(GLUE) benchmark \citep{wang2018glue} \cjy{covers diverse NLU} tasks, which is the main benchmark used in \cjy{PLMs}. To explore whether synonym knowledge deteriorates performance on \cjy{general NLU} tasks, we evaluate PICSO on eight datasets of GLUE, \cjy{and the results} are shown in Table \ref{tab:glue}.

\cjy{In summary, PICSO achieves competitive results on GLUE, and some interesting phenomena can be observed.} 
(1) PICSO has the highest average score \cjy{on all the 8} tasks, which proves that \cjy{the} synonym knowledge is \cjy{beneficial. SAPBERT, on the other hand, struggles to handle the context-involved tasks, achieving the worst result.}
(2) \cjy{These GLUE tasks mainly have two categories}. The first category \cjy{involves similarity prediction}, which includes MRPC, STS-B and QQP. \cjy{They require the model to infer whether two sentences have paraphrase/semantic equivalence}. PICSO and LIBERT achieved the best results \cjy{as} synonym knowledge is particularly valuable.
(3) \cjy{The second} category is natural language inference tasks \cjy{including} MNLI, QNLI and RTE. The goal of these tasks is to determine whether \cjy{two sentences have} implicative relations (i.e., whether the hypothesis can be inferred from the premises), which requires factual knowledge. \cjy{PLMs with KG knowledge injected, i.e., K-Adapter, achieve better} performance.

\subsection{Additional Experiments}
\subsubsection{Ablation Study on Entity-Aware Adapter}
\label{exp:abla_entity_aware}
In the ablation experiments, for the similarity-oriented tasks, we selected entity linking and lexical simplification as representatives of the fine-tuned and unsupervised methods, respectively. For the NLU tasks, STS-B and QQP are picked. 

We conducted ablation experiment for the entity-aware mechanism to demonstrate its effectiveness and necessity for injecting contextual synonym knowledge. The results in Table \ref{tab:abla_entity_aware} show that although PICSO w/o EA does not show the same negative gain as the SAPBERT and LIBERT, the magnitude of the gain is much smaller compared to PICSO, especially for the unsupervised lexical simplification. We argue that with entity-aware Adapter, the model can focus more on the semantics represented by the entities and thus more accurately draw the corresponding synonyms closer without introducing noise.

\begin{table}[ht]
\centering
\caption{Ablation study on entity-aware Adapter. w/o EA means without entity-aware mechanism.}
\scalebox{0.72}{
\renewcommand\arraystretch{1}
\setlength\tabcolsep{3.5pt}
\begin{tabular}{l|cc|cccc|cc}
\toprule
\multirow{3}{*}{Model}   & \multicolumn{2}{c|}{\textbf{Entity Linking}} & \multicolumn{4}{c|}{\textbf{Lexical Simplification}} & \multicolumn{2}{c}{\textbf{GLUE}} \\ \cmidrule{2-9} 
                         & \multicolumn{2}{c|}{\textbf{AIDA}}           & \multicolumn{4}{c|}{\textbf{BenchLS}}                & \textbf{STS-B}      & \textbf{QQP}         \\ \cmidrule{2-9} 
                         & \textbf{Acc@1}            & \textbf{Acc@5}            & \textbf{P}         & \textbf{R}         & \textbf{F1}       & \textbf{Acc}      & \textbf{Pearson}        & \textbf{F1}     \\ \midrule
BERT                & 72.35            & 86.64           & 24.81     & 32.08    & 27.98    & 57.03    & 88.48     & 87.49
\\ \midrule\midrule
PICSO                & 78.26            & 88.63           & 25.18     & 34.18    & 29.00    & 59.63    & 89.43     & 88.58         \\ \midrule
w/o EA & 73.23            & 87.08           & 24.80     & 32.89    & 28.27    & 57.68    & 88.96     & 87.54       \\ \bottomrule
\end{tabular}}
\label{tab:abla_entity_aware}
\end{table}

\subsubsection{Comparison of Pre-training Loss Functions}
\label{sec:loss}
As illustrated in Table \ref{tab:loss}, we compare the effects of three pre-training objectives on the pre-training effect. Triplet Margin Loss \citep{balntas2016learning} was designed for computer vision tasks such as image classification and can be formulated as $L=\left[d_{a p}-d_{a n}+m\right]_{+}$, where $d_{a p}$ and $d_{a n}$ denote the distance from the anchor to the positive and negative samples, respectively. InfoNCE Loss is a classic contrastive loss and has been widely used in self-supervision papers \citep{he2020momentum,chen2020simple}. It can be equated as $L=-\log \frac{\exp \left(q \cdot k_{+} / \tau\right)}{\sum_{i=0}^K \exp \left(q \cdot k_i / \tau\right)}$, in which $q$, $k_+$, $k_i$ represent anchor, positive and negative samples respectively. Note that online hard negatives mining is available for both pre-training objectives. In contrast, our designed pre-training objective does not require cumbersome online mining, but only re-weights the penalty for negative samples based on similarity, thus allowing the model to focus more on non-synonymous entity pairs that are elusive to distinguish. Our proposed pre-training objective is more efficient and effective.

\begin{table}[ht]
\centering
\caption{Comparison of different pre-training loss functions.}
\scalebox{0.72}{
\renewcommand\arraystretch{1}
\setlength\tabcolsep{3.5pt}
\begin{tabular}{l|cc|cccc|cc}
\toprule
\multirow{3}{*}{Loss}   & \multicolumn{2}{c|}{\textbf{Entity Linking}} & \multicolumn{4}{c|}{\textbf{Lexical Simplification}} & \multicolumn{2}{c}{\textbf{GLUE}} \\ \cmidrule{2-9} 
                         & \multicolumn{2}{c|}{\textbf{AIDA}}           & \multicolumn{4}{c|}{\textbf{BenchLS}}                & \textbf{STS-B}      & \textbf{QQP}         \\ \cmidrule{2-9} 
                         & \textbf{Acc@1}            & \textbf{Acc@5}            & \textbf{P}         & \textbf{R}         & \textbf{F1}       & \textbf{Acc}      & \textbf{Pearson}        & \textbf{F1}     \\ \midrule
Triplet Margin Loss                & 77.23            & 87.12           & 23.69     & 32.18    & 27.29    & 54.76    & 87.28     & 85.73        \\ \midrule
InfoNCE                & 77.32            & 87.32           & 23.46     & 32.10    & 27.11    & 54.27    & 88.05     & 85.71        \\ \midrule
Ours & 78.26            & 88.63           & 25.18     & 34.18    & 29.00    & 59.63    & 89.43     & 88.58        \\ \bottomrule
\end{tabular}}
\label{tab:loss}
\end{table}

\subsubsection{Comparison of PLM backbones}
In the main experiments we adopt BERT as the backbone of PICSO. Roberta differs from BERT in that it discards next sentence prediction (NSP) as the pre-training task. In Table \ref{tab:exp_backbone}, we compare the performances of PICSO with BERT and Roberta as the backbone, and their average gains on downstream tasks are 2.24 and 2.14, respectively. Benefiting from the favorable performance of Roberta on the entity linking task, Roberta-based PICSO outperforms BERT-based PICSO on the entity linking task. We can conclude that the enhancement of contextual synonym knowledge for downstream tasks is model-agnostic and universal.

\begin{table}[h]
\centering
\caption{Comparison of different PLM backbones.}
\scalebox{0.72}{
\renewcommand\arraystretch{1}
\setlength\tabcolsep{3.5pt}
\begin{tabular}{l|cc|cccc|cc}
\toprule
\multirow{3}{*}{Model}   & \multicolumn{2}{c|}{\textbf{Entity Linking}} & \multicolumn{4}{c|}{\textbf{Lexical Simplification}} & \multicolumn{2}{c}{\textbf{GLUE}} \\ \cmidrule{2-9} 
                         & \multicolumn{2}{c|}{\textbf{AIDA}}           & \multicolumn{4}{c|}{\textbf{BenchLS}}                & \textbf{STS-B}      & \textbf{QQP}         \\ \cmidrule{2-9} 
                         & \textbf{Acc@1}            & \textbf{Acc@5}            & \textbf{P}         & \textbf{R}         & \textbf{F1}       & \textbf{Acc}      & \textbf{Pearson}        & \textbf{F1}     \\ \midrule
BERT                & 72.35            & 86.64           & 24.81     & 32.08    & 27.98    & 57.03    & 88.48     & 87.49        \\ \midrule
Roberta                & 74.89            & 87.32           & 20.08     & 27.25    & 23.12    & 52.41    & 84.88     & 87.48        \\ \midrule\midrule
PICSO                & 78.26            & 88.63           & 25.18     & 34.18    & 29.00    & 59.63    & 89.43     & 88.58        \\ \midrule
Roberta-based & 78.34            & 89.10           & 21.38     & 28.91    & 24.58  & 54.84    & 87.13     & 88.89       \\ \bottomrule
\end{tabular}}
\label{tab:exp_backbone}
\end{table}

\subsubsection{Case Study}
Case study is conducted on four similarity-oriented tasks. From Table \ref{tab:case}, we can observe that: (1) The example on the entity linking task illustrates that BERT struggles to resolve the ambiguity caused by multiple meanings of a word, while benefiting from the introduction of contextual information about synonyms, PICSO can clearly distinguish non-synonymous entity pairs with the same surface name. This is more obviously illustrated by the example on the lexical simplification task, where tender has the meaning of gentle in some contexts, and painful when describing a body part. Compared to BERT, PICSO can distinguish more explicitly between the two semantics. (2) The case on entity resolution and KG canonicalization demonstrates that BERT lacks the ability to capture synonym information. For example, \textit{Sky Caption Blue} and \textit{Blue} are synonyms in some sense. Whereas on the KG canonicalization task, BERT fails to infer the exact synonym pair and tended to consider \textit{Virginia wesleyan college} and \textit{Rider college} as synonyms, which contain the common token \textit{college}, or consider \textit{Palm casino} and \textit{Flamingo hotel} as synonyms, which are both buildings in Las Vegas.

\begin{table*}[h]
\centering
\caption{Case study for similarity-oriented tasks.}
\scalebox{0.9}{
\begin{tabular}{lp{5cm}p{5cm}p{4cm}}
\toprule
\multirow{2}{*}{\textbf{Task}}                 & \multicolumn{2}{c}{\textbf{Output of}}                            & \multicolumn{1}{c}{\multirow{2}{*}{\textbf{Ground Truth}}} \\ \cmidrule{2-3}
                                      & \multicolumn{1}{c}{\textbf{PICSO}} & \multicolumn{1}{c}{\textbf{BERT}} & \multicolumn{1}{c}{}                              \\ \midrule
Entity Linking       & \textbf{Mention}: ``or 10 continues west into  \textit{\color{blue}{beaverton}}, where it interchanges with oregon route 217, a freeway.'' \newline \textbf{Ranked Candidate}: [``\textbf{beaverton, oregon}'', ``beaverton'', ``beaverton, ontario'']                             &  \textbf{Mention}: ``or 10 continues west into  \textit{\color{blue}{beaverton}}, where it interchanges with oregon route 217, a freeway.''\newline  \textbf{Ranked Candidate}: [``beaverton'', ``\textbf{beaverton, oregon}'', ``beaverton, ontario'']                        &                           beaverton, oregon                                                 \\ \midrule
Entity Resolution    & \textbf{Entity 1}: ``TomTom Runner 2 Cardio+Music DBL/LBL (Large) - \textit{Sky Captain Blue/Scuba Blue} TomTom Running Accessories Blue''\newline \textbf{Entity 2}: ``TomTom Runner 2 Cardio GPS Watch with Music Large Strap - \textit{Blue Blue }''\newline \textbf{Prediction}: Same                               &   \textbf{Entity 1}: ``TomTom Runner 2 Cardio+Music DBL/LBL (Large) - \textit{Sky Captain Blue/Scuba Blue} TomTom Running Accessories Blue''\newline \textbf{Entity 2}: ``TomTom Runner 2 Cardio GPS Watch with Music Large Strap - \textit{Blue Blue }''\newline \textbf{Prediction}: Not Same & Same                                                                           \\ \midrule
Lexical Simplification   & \textbf{Sentence}: Women usually notice little change in their breasts, but if you are a man, your breasts may become slightly larger and may be \textit{\color{blue}{tender}}. \newline \textbf{Candidate:} [strong, gentle, serious, \textbf{sensitive}, weak, \textbf{painful}]  &  \textbf{Sentence}: Women usually notice little change in their breasts, but if you are a man, your breasts may become slightly larger and may be \textit{\color{blue}{tender}}. \newline \textbf{Candidate:} [strong, gentle, soft, special, weak, sweet]                        & [sore, sensitive, painful]                                                                           \\ \midrule
\multirow{2}{*}{KG  Canonicalization} &  \{Virginia wesleyan, Virginia wesleyan college \}                            &  \{Virginia wesleyan college, \textit{\color{red}{Columbus college}}, \textit{\color{red}{Rider college}} \}                         &  \{Virginia wesleyan, Virginia wesleyan college \}                                                 \\ \cmidrule{2-4} 
                                      &  \{Flamingo hotel, Flamingo la vega \}                             &  \{\textit{\color{red}{Palm casino}}, Flamingo hotel \}                        & \{Flamingo hotel, Flamingo la vega\}                                                  \\ \bottomrule
\end{tabular}}
\label{tab:case}
\end{table*}

\section{Conclusion and Future Work}
The paper presents \cjy{PICSO that can inject contextual synonym knowledge from multiple domains into the PLM without disrupting its original semantic understanding capabilities.
PICSO are equipped with entity-aware Adapters, each of which constrains the visible range of the tokens of the synonyms through a masked self-attention mechanism for learning the semantics of the entity and its context.
With the contextual synonym knowledge from Wikidata  (general domain) and UMLS (medical domain), PICSO often dramatically outperforms the original PLMs and the other knowledge and synonym injection PLMs on four different similarity-oriented tasks, and can also  benefit general NLU tasks in GLUE.}
In the future, we will investigate a multi-task pre-training paradigm for synonym knowledge \cjy{injection} to better exploit the synonyms widely available in \cjy{both} unstructured text and structured KGs, \cjy{and will evaluate our methods on more similarity-oriented tasks, e.g. ontology alignment.}

\section*{Acknowledgement}
This research is supported by National Natural Science Foundation of China (Grant No.62276154 and 62011540405), Beijing Academy of Artificial Intelligence (BAAI), the Natural Science Foundation of Guangdong Province (Grant No. 2021A1515012640), Basic Research Fund of Shenzhen City (Grant No. JCYJ20210324120012033), and Overseas Cooperation Research Fund of Tsinghua Shenzhen International Graduate School  (Grant No. HW2021008).

\printcredits

\bibliographystyle{cas-model2-names}

\bibliography{cas-refs}

\bio{Figures/liyangning}
Yangning Li received the BEng degree from the Department of Computer Science and Technology, Huazhong University of Science and Technology, in 2020. He is currently working toward a Master's degree with the Tsinghua Shenzhen International Graduate School, Tsinghua University. His research interests include natural language processing and data mining.
\endbio
\newpage

\bio{Figures/jiaoyanchen}
Jiaoyan Chen is permanent lecturer at Department of Computer Science, The University of Manchester. 
He used to be a Senior Researcher at Department of Computer Science, University of Oxford, and got his Ph.D and Bachelor degree in Computer Science and Technology in Zhejiang University. 
His research interests includes knowledge graph, ontology, symbolic and sub-symbolic reasoning, neural-symbolic AI, etc.
\endbio

\bio{Figures/liyinghui}
Yinghui Li received the BEng degree from the Department of Computer Science and Technology, Tsinghua University, in 2020. He is currently working toward the PhD degree with the Tsinghua Shenzhen International Graduate School, Tsinghua University. His research interests include natural language processing and deep learning.
\endbio
\vspace{1cm}

\bio{Figures/tianyu}
Tianyu Yu received the bachelor's degree in software engineering from the Beihang University, CHina. He is currently a graducate student major in computer technology in Tsinghua University. His research interest include information extraction, knowledge representation learning and cross-modal representation learning.
\endbio
\vspace{1cm}

\bio{Figures/chenxi}
Xi Chen received his PhD degree in computer science from the Zhejiang University. He is currently the head of the cross-modal algorithm center of Tencent Platform and Content Group and mainly focuses on various applications of NLP.
\endbio
\vspace{1cm}

\bio{Figures/zhenghaitao}
Hai-Tao Zheng received the bachelor’s and master’s degrees in computer science from the Sun Yat-Sen University, China, and the PhD degree in medical informatics from Seoul National University, South Korea. He is currently an associate professor with the Shenzhen International Graduate School, Tsinghua University, and also with Peng Cheng Laboratory. His research interests include web science, semantic web, information retrieval, and machine learning.
\endbio
\end{document}